\documentclass[letter, 10 pt, conference]{ieeeconf}  

\IEEEoverridecommandlockouts                              

\usepackage{amsmath}
\usepackage{amssymb}
\usepackage{mathtools}
\usepackage{graphicx}
\usepackage{float}
\usepackage{array}
\usepackage{tikz}
\usepackage{listings}
\usepackage{hyperref}
\usepackage{graphicx}
\usepackage{cite}
\usepackage{dsfont}
\usepackage{mathtools,etoolbox}
\usepackage[ruled, linesnumbered]{algorithm2e}
\usepackage[none]{hyphenat}
\usepackage[utf8]{inputenc}
\usepackage[english]{babel}
\usepackage[T1]{fontenc}

\usepackage{amsthm}
\usepackage{xfrac}
\usepackage{nicefrac}
\usepackage{booktabs}
\usepackage{flushend}
\usepackage[switch, pagewise]{lineno}
\usepackage{mathrsfs}
\DeclareMathAlphabet{\mathpzc}{OT1}{pzc}{m}{it}
\usepackage{multirow}
\usepackage{multicol}
\usepackage{bbm}
\usepackage{soul}
\usepackage{xfrac}
\usepackage{balance}
\usepackage[final]{pdfpages}

\DeclareMathOperator*{\argmin}{argmin} 
\DeclareMathOperator*{\argmax}{argmax} 

\hypersetup{colorlinks=true,urlcolor=blue,citecolor=red}
\usepackage[some, placement=top]{background}

\backgroundsetup{
        placement=top,
        position={8cm,2cm}}

\SetBgScale{1}
\SetBgContents{\parbox{16cm}{\Large \centering This paper has been accepted for publication at the IEEE International Conference on Robotics and Automation (ICRA),  Paris, 2020. \textcopyright IEEE}}
\SetBgColor{gray}
\SetBgAngle{0}
\SetBgOpacity{0.9}

\title{\LARGE \bf EVDodgeNet: Deep Dynamic Obstacle Dodging with Event Cameras}

\author{Nitin J. Sanket$^{1}$, Chethan M. Parameshwara$^{1}$, Chahat Deep Singh$^{1}$, Ashwin V. Kuruttukulam$^{1}$,\\ Cornelia Ferm{\"u}ller$^{1}$, Davide Scaramuzza$^{2}$, Yiannis Aloimonos$^{1}$ 
\thanks{\textit{Nitin J. Sanket and Chethan M. Parameshwara
 contributed equally to this work. (Corresponding author: Nitin
J. Sanket.)}}
\thanks{$^{1}$Perception and Robotics Group, University of Maryland Institute for Advanced Computer Studies, University of Maryland, College Park.}
\thanks{$^{2}$Robotics and Perception Group, Dep. of Informatics,
University of Zurich, and Dep. of Neuroinformatics, University of Zurich and ETH Zurich.}} 

\begin{document}

\maketitle
\thispagestyle{plain}
\pagestyle{plain}
\BgThispage


\begin{abstract}

Dynamic obstacle avoidance on quadrotors requires low latency. A class of sensors that are particularly suitable for such scenarios are event cameras. In this paper, we present a deep learning based solution for dodging multiple dynamic obstacles on a quadrotor with a single event camera and on-board computation. Our approach uses a series of shallow neural networks for estimating both the ego-motion and the motion of independently moving objects.
The networks are trained in simulation and directly transfer to the real world without any fine-tuning or retraining.
We successfully evaluate and demonstrate the proposed approach in many real-world experiments with obstacles of different shapes and sizes, achieving an overall success rate of 70\% including objects of unknown shape and a low light testing scenario. To our knowledge, this is the first deep learning -- based solution to the problem of dynamic obstacle avoidance using event cameras on a quadrotor. Finally, we also extend our work to the pursuit task by merely reversing the control policy, proving that our navigation stack can cater to different scenarios.

\end{abstract}


\section*{Supplementary Material}
The accompanying video, supplementary material, code and dataset are available at
\url{http://prg.cs.umd.edu/EVDodgeNet} 



\section{Introduction and Philosophy}

The never-ending quest to understand and mimic ultra-efficient flying agents like bees, flies, and birds has fueled the human fascination to create autonomous, agile and ultra-efficient small aerial robots. These robots are not only utilitarian but are much safer to operate in static or dynamic environments and around other agents as compared to their larger counterparts. Need for creation of such small aerial robots has given rise to the development of numerous perception algorithms for low latency obstacle avoidance. Here, latency is defined as the time the robot takes to perceive, interpret and generate control commands \cite{howfastistoofast}.

Low latency static obstacle avoidance has been studied extensively in the last decade \cite{sermanet2007speed}. Recently, however, dynamic obstacle avoidance has gained popularity in the field of robotics due to the exponential growth of event cameras.
These are bioinspired vision sensors that output per-pixel temporal intensity differences caused by relative motion with microsecond latency~\cite{eventsurvey}.

Event cameras have the potential to become the de-facto standard for visual motion estimation problems due to their inherent advantages of low latency, high temporal resolution, and high dynamic range~\cite{ultimateslam}. These advantages make event cameras tailor made for dynamic obstacle avoidance.

\begin{figure}[t!]
    \centering
    \includegraphics[width=\columnwidth]{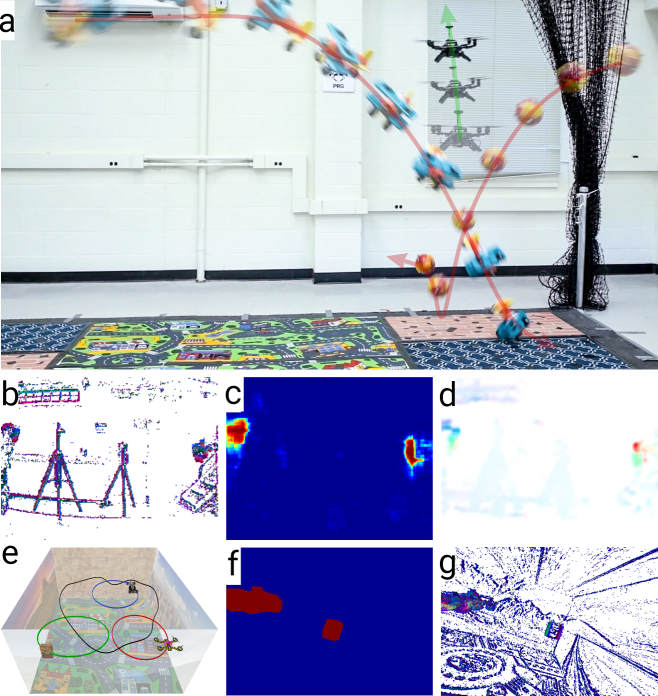}
    \caption{(a) A real quadrotor running \textit{EVDodgeNet} to dodge two obstacles thrown at it simultaneously. (b) Raw event frame as seen from the front event camera. (c) Segmentation output. (d) Segmentation flow output which includes both segmentation and optical flow. (e) Simulation environment where \textit{EVDodgeNet} was trained. (f) Segmentation ground truth. (g) Simulated front facing event frame. \textit{All the images in this paper are best viewed in color.}}
    \label{fig:Banner}
\end{figure}

In this paper, we present a framework to dodge multiple unknown dynamic obstacles on a quadrotor with event cameras using deep learning. Although dynamic obstacle detection using traditional cameras and deep learning has been extensively studied in the computer vision community under the umbrellas of object segmentation and detection, they are either of high latency, computationally expensive (not enough to be used on micro/nano-quadrotors) and/or do not generalize to novel objects without retraining or fine-tuning.

Our work is closely related to \cite{howfastistoofast} with the key difference being that our approach uses deep learning and generalizes to unknown real objects after being trained only on simulation.

\subsection{Problem Formulation and Contributions}
A quadrotor moves in a static scene with multiple Independently Moving dynamic Objects/obstacles (IMOs). The quadrotor is equipped with a front facing event camera, a downfacing lower resolution event camera coupled with sonar, for altitude measurements, and an IMU.

The problem we address is as follows: \textit{Can we present an AI framework for the task of dodging/evading/avoiding these dynamic obstacles without any prior knowledge, using only on-board sensing and computation?}

We present various flavors of the dodging problem, such as hovering quadrotor dodging unknown obstacles, slow-moving quadrotor dodging unknown shaped obstacles given a bound on size, hovering and slow moving quadrotor dodging known objects (particularly targeted to spherical objects of known radii). We extend our approach by demonstrating pursuit/intercept of a known object using the same deep-learning framework. This showcases that our proposed framework can be used in a general navigation stack on a quadrotor and can be re-purposed for various related tasks. A summary of our contributions are (Fig. \ref{fig:Banner}):
\begin{itemize}
\item We propose and implement a network (called EVDeBlurNet) that \textit{deblurs} event frames, such that learning algorithms trained on simulated data can generalize to real scenes without retraining or fine-tuning.
\item We design and implement a network (called EVSegFlowNet) that performs both segmentation and optical flow of IMOs to obtain both segmentation and optical flow in a single network.
\item We propose a control policy based on estimated motion of multiple IMOs under various scenarios.
\item{We evaluate and demonstrate the proposed approach on a real quadrotor with onboard perception and computation.}
\end{itemize}

\subsection{Related Work}
We subdivide the related work into three parts, i.e., ego-motion estimation, independent motion segmentation, and obstacle avoidance.

\subsubsection{Independent Motion Detection and Ego-motion Estimation -- Two sides of the same coin}
Information from complementary sensors, such as standard cameras and Inertial Measurement Units (IMUs), has given rise to the field of Visual Inertial Odometry (VIO) \cite{ROVIO, VINS}. Low latency VIO algorithms based on event cameras have been presented in \cite{evio, ultimateslam}, which use classical feature tracking inspired methods to estimate ego-motion. Other works, instead, try to add semantic information to enhance the quality of odometry
by adding strong priors about the scene \cite{salientdso, kostas-semantic-SLAM}. Most works in the literature focus on ego-motion estimation in static scenes which are seldom encountered in the real world. To account for moving objects, these algorithms implement a set of outlier rejection schemes to detect IMOs. We would like to point out that by carefully modelling these ``outliers'' one can estimate both ego-motion and IMO motion~\cite{sabzevariTRO16}. 

\subsubsection{Image stabilization as a key to independent motion segmentation}
Keen readers might have contrived that by performing the process of image stabilization IMOs would ``stand-out''. Indeed, this was the approach most robust algorithms used in the last two decades. A similar concept was adapted in some recent works on event-based cameras for detecting IMOs \cite{vasco2017independent, AntonEVTracker, stoffregen2019event}. Recently a deep learning based approach was presented for IMO detection using a structure from motion inspired approach  \cite{evimo}. 

\subsubsection{Obstacle avoidance on aerial robots}
The works presented in the above two subsections have aided the advancement of obstacle avoidance on aerial robots. \cite{alvarez2016collision, gapflyt} presented approaches for high speed static obstacle avoidance by estimating depth maps and visual servoing using a monocular imaging camera respectively. \cite{barry2018high}  provides a detailed collation of the prior work on static obstacle avoidance using stereo cameras.  A hardware and software architecture was proposed in \cite{mohta2018fast, mohta2018experiments} for high speed quadrotor navigation by mapping the cluttered environment using a lidar. Using event cameras for high speed dodging is not new and the first work was presented in \cite{mueggler2015towards} where an approach was presented to avoid a dynamic spherical obstacle using stereo event cameras. Very recently, \cite{howfastistoofast} presented a detailed analysis of perception latency for dodging a dynamic obstacle.

\subsection{Organization of the paper}
\label{subsec:organization}
The paper is structured into perception and control modules. The perception module (Refer to Fig. \ref{fig:Overview}) is further divided into three segments.\\
1. The input to the perception system are event frames (Sec. \ref{subsec:db}). Such a projection of event data to generate \textit{event frames} suffers from misalignment \cite{ContrastMaximization} unless motion compensation is performed. We call this misalignment or loss of contrast/sharpness as \textit{blur} due to its visual resemblance to classical image motion blur. To perform motion compensation and denoising, we present a neural network called \textit{EVDeBlurNet} in Sec. \ref{subsec:db}.\\
2. Suppl. Sec. {\color{red}S.III.} presents how ego-motion is obtained using \textit{EVHomographyNet}. \\
3. Sec. \ref{subsec:segflownet} describes how segmentation and optical flow of IMOs are obtained using the novel \textit{EVSegFlowNet}.\\
Sec. \ref{sec:Evasion} presents the control scheme for dodging given the outputs from the perception module. We also bring the generality of our perception stack into limelight in Suppl. Sec. {\color{red}S.IX} by adapting our approach to the problem of pursuit. Sec. \ref{sec:Expts} illustrates the experimental setup and provides error analyses of the approaches presented along with detailed ablation studies. We finally conclude the paper in Sec. \ref{sec:Conc} with parting thoughts on future work.

\section{Deep Learning Based Navigation Stack For Dodging Dynamic Obstacles}
\begin{figure}[t!]
    \centering
    \includegraphics[width=\columnwidth]{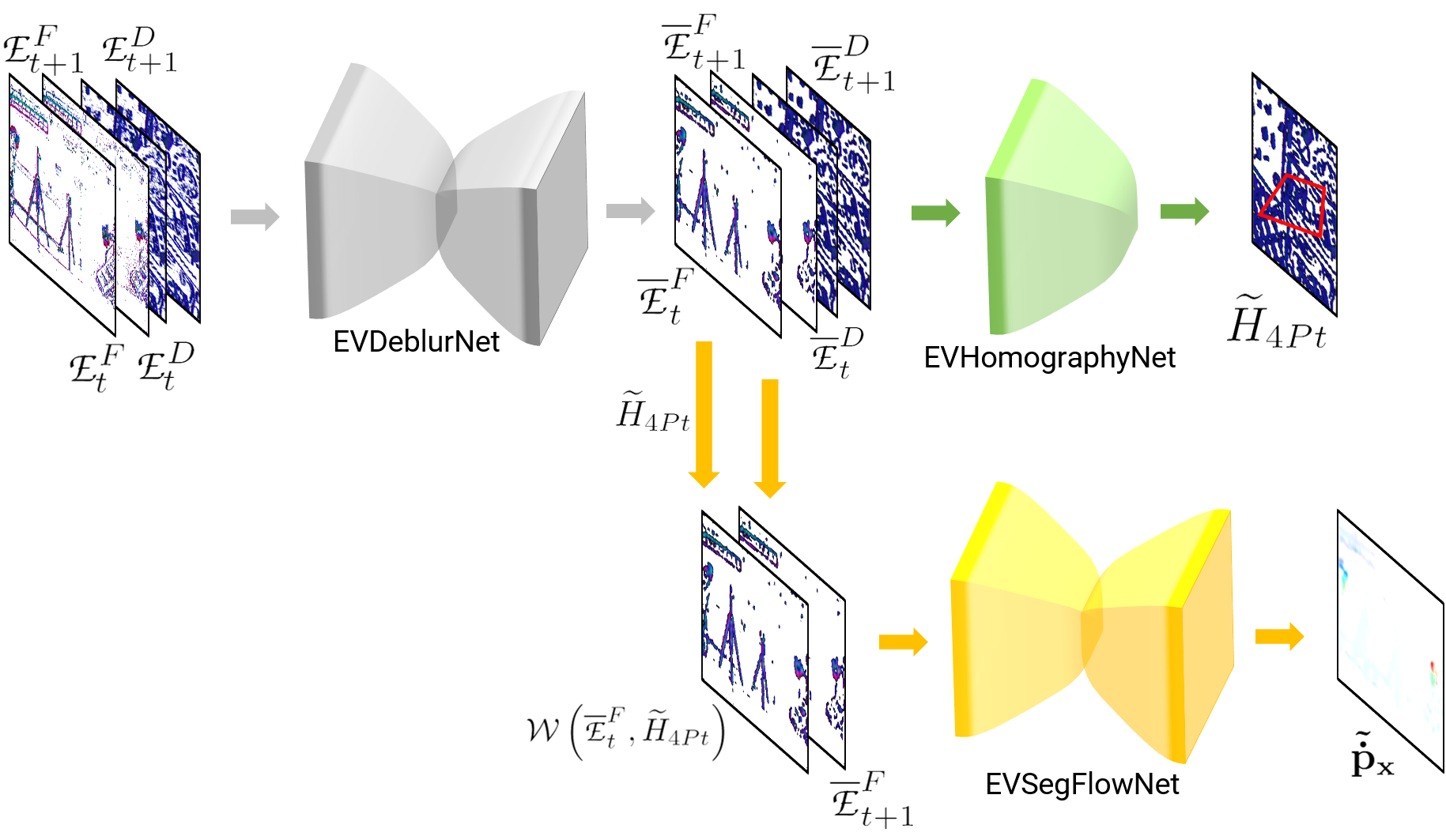}
    \caption{Overview of the proposed neural network based navigation stack for the purpose of dodging.}
    \label{fig:Overview}
\end{figure}

\begin{figure}
     \centering
     \includegraphics[width=\columnwidth]{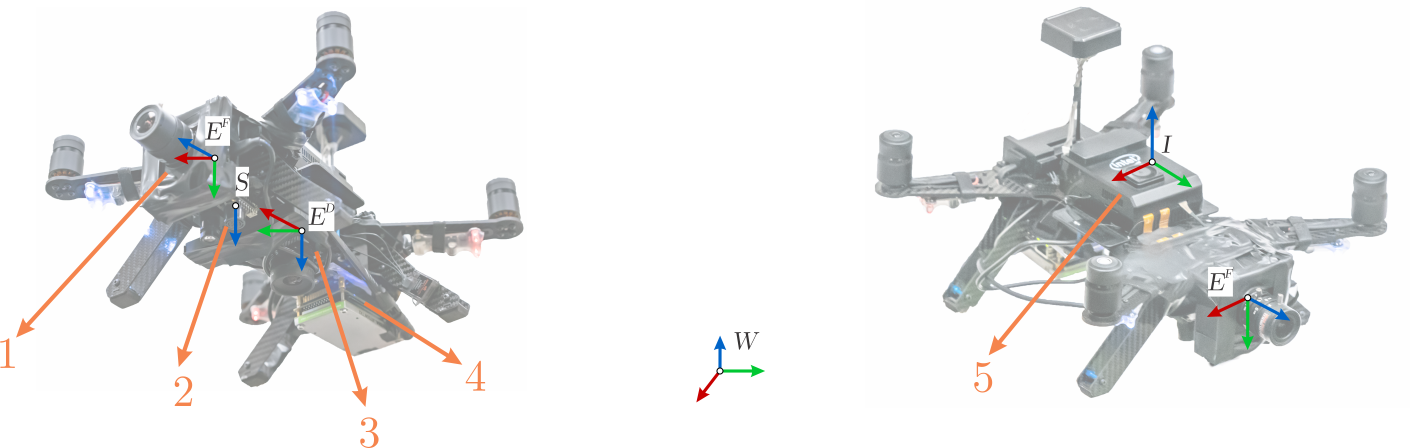}
     \caption{Representation of coordinate frames on the hardware platform used. (1) Front facing DAVIS 240C, (2) down facing sonar on PX4Flow, (3) down facing DAVIS 240B, (4) NVIDIA TX2 CPU+GPU, (5) Intel$^\text{\textregistered}$ Aero Compute board.}
     \label{fig:CoordinateFrames}
 \end{figure}

Fig. \ref{fig:Overview} shows an overview of our proposed approach. Refer to Suppl. Sec. {\color{red}S.I} for the coordinate frame definitions. Our hardware setup is shown in Fig. \ref{fig:CoordinateFrames}.

\subsection{EVDeBlurNet}
\label{subsec:db}
The event frame  $\mathpzc{E}$ consists of three channels. The first and second channels contain the per-pixel average count of positive and negative events. The third channel contains the per-pixel average time between events (refer to Sec. {\color{red}S.II} for a mathematical formulation). Though event representation offers many advantages regarding computational complexity and providing tight time bounds on operation, there is a hitch.  
Event frames can be ``blurry'' (projection of misaligned events) based on a combination of the integration time $\delta t$ (observe in Fig. \ref{fig:DBOutputs} how sharpness of the image decreases as integration time $\delta t$ increases), apparent scene movement on the image plane (which depends on the amount of camera movement and depth of the scene) and scene contrast (contrast of the latent image pixels). Here, we define blur on the event frame $\mathpzc{E}$ as the misalignment of the events in a small integration time $\delta t$.

An event is triggered when the relative contrast (on the latent image $I$) exceeds a threshold $\tau$ and is mathematically modelled as: $\Vert \log\left(I\right) \Vert_1 \approx \Vert \langle \nabla_{\mathbf{x}} \log\left(I \right), \mathbf{\dot{x}}\Delta t\rangle \Vert_1 \ge \tau$.


Here,  $\langle \cdot, \cdot \rangle$ denotes the inner/dot product between two vectors, $\nabla_{\mathbf{x}}$ is the spatial gradient, $\mathbf{\dot{x}}$ is the motion field on the image and $\Delta t$ is the time since the previous event at the same location. The above equation elucidates how the latent image contrast, motion and depth are coupled to event frames. Note that, $\mathbf{\dot{x}}$ depends on the 3D camera motion and the scene depth. We refer the reader to \cite{gallego2015event} for more details.

This ``blurriness" of the event frame can adversely affect the performance of algorithms built on them. To alleviate this problem, we need to deblur the event images. This is fairly easy if we directly use the spatio-temporal event cloud and follow the approach described in \cite{ContrastMaximization}. Essentially the problem deals with finding point trajectories along the spatio-temporal point cloud to maximize a heuristically chosen contrast function. Mathematically, we want to solve the following problem: $\argmax_{\theta}\,\, \mathcal{C}\left(\mathcal{W}\left(\mathpzc{E}, \theta \right)\right)$. Here $\mathcal{C}$ is a heuristic contrast function and $\theta$ are the parameters of point trajectories in the spatio-temporal point cloud according to which the events are warped and $\mathcal{W}\left(\mathpzc{E}, \theta \right)$ represents the event image formed by the warped events. In our scenario, we want to model the deblurring problem in 2D, i.e., working on $\mathpzc{E}$ directly without the use of a spatio-temporal point cloud so that the problem can be solved efficiently using a  2D Convolutional Neural Network (CNN). Such a deblurring problem using a single image has been studied extensively for traditional cameras for rectifying motion blurred photos. Our modified problem in 2D can be formulated as: $\argmax_{\mathcal{K}}\,\, \mathcal{C}\left(\mathcal{K}\circledast\mathpzc{E}\right)$. Here $\mathcal{K}$ is the heterogeneous deblur kernel and $\circledast$ is the convolution operator. However, estimating $\mathcal{K}$ directly is not constrained enough to be learned in an unsupervised manner. Instead, we formulate the deblurring problem inspired by Total Variation (TV) denoising to give us the final optimization problem as follows: $\argmax_{\mathpzc{\overline{E}}}\,\,  \mathcal{C}\left(\mathpzc{\overline{E}}\right) + \lambda \argmin_{\mathpzc{\overline{E}}}\mathcal{D}\left(\mathpzc{E}, \mathpzc{\overline{E}}\right)$. Here $\mathpzc{\overline{E}}$ represents the deblurred event frame, $\lambda$ is a regularization penalty and $\mathcal{D}$ represents a distance function to measure similarity between two event frames. Note that directly solving $\argmax_{\mathpzc{\overline{E}}}\,\,  \mathcal{C}\left(\mathpzc{\overline{E}}\right)$ yields trivial solutions of high frequency noise.

To learn the function using a neural network we convert the $\argmax$ operator into an $\argmin$ operator as follows:
\begin{equation} \argmin_{\mathpzc{\overline{E}}}\,\,  -\mathcal{C}\left(\mathpzc{\overline{E}}\right) + \lambda \mathcal{D}\left(\mathpzc{E}, \mathpzc{\overline{E}}\right)\end{equation}
Refer to Suppl. Sec. {\color{red}{III}} for a detailed mathematical description of all the loss functions.
Intuitively, the higher the value of the contrast, the lower the value of the loss function, but going away too far from the input will penalize the loss function striking a balance between high contrast and similarity to the input image. We call our CNN which generates the deblurred event images \textit{EVDeBlurNet}. It takes as input $\mathpzc{E}$ and outputs deblurred $\mathpzc{\overline{E}}$. The network architecture is a simple encoder-decoder with four convolutional and four deconvolutional layers with batch normalization (Suppl. Sec. {\color{red}S.V}). Another benefit of the encoder decoder's lossy reconstruction is that it removes stray events (which are generally noise) and retains events corresponding to contours, thereby greatly increasing the signal to noise ratio.

Recently, \cite{zhu2019unsupervised} also presented a method for deblurring event frames to improve optical flow estimation via a coupling between predicted optical flow and sharpness in the event frame in the loss function. In contrast, our work presents a problem-independent deblurring network without the supervision from optical flow. We obtain ``cheap'' odometry using the \textit{EVHomographyNet} as described in Suppl. Sec. {\color{red}S.III} which is built upon \cite{deepHomography, unsupHomography}.

\subsection{EVSegFlowNet}
\label{subsec:segflownet}
The end goal of this work is to detect/segment Independently Moving Objects (IMOs) and to dodge them. One could fragment this problem into two major parts, detecting IMOs, and subsequently estimating their motion to issue a control command to move away from the IMO in a safe manner. Let's start by discussing each fragment.
Firstly, we want to segment the object using consecutive event frames $\mathpzc{E}_t$ and $\mathpzc{E}_{t+1}$. A simple way to accomplish this is by generating simulated data with known segmentation for each frame and then training a CNN to predict the foreground (IMO)/background segmentation. Such a CNN can be trained using a simple cross-entropy loss function: $ \argmin_{p_{f}}\,\, -\mathbb{E}\left( \mathds{1}_f\log\left(p_f\right) + \mathds{1}_b\log\left(p_b\right)\right)$. Here, $\mathds{1}_f, \mathds{1}_b$ are the indicator variables denoting if a pixel belongs to foreground or background. They are mutually exclusive, i.e., $\mathds{1}_f = \lnot \mathds{1}_b$ and $p_f, p_b$ represent the foreground and background predicted probabilities where $p_f + p_b = 1$. Note that each operation in the above equation is performed per pixel, and then an average over all pixels is computed.
In the second step we want to estimate the IMO motion. Without any prior knowledge about the IMO it is impossible to estimate the 3D motion of the IMO from a monocular camera (event based or traditional). To make this problem tractable, we assume a prior about the object. More details can be found in Sec. \ref{sec:Evasion}.





Once we have a prior about the object, we can estimate the 3D IMO motion using optical flow of the pixels corresponding to the IMO on the image plane. A simple way to obtain optical flow is to train a CNN in a supervised manner. However, recent research has shown that these do not generalize well to new scenes/objects \cite{meister2018unflow}. A better way is to use a self-supervised or completely unsupervised loss function: $\argmin_{\mathbf{\dot{x}}}\,\, \mathbb{E}\left(\mathcal{D}\left(\mathcal{W} \left(\mathpzc{E}_{t}, \mathbf{\dot{x}}\right), \mathpzc{E}_{t+1}\right) \right)$. Here $\mathbf{\dot{x}}$ is the estimated optical flow between $\mathpzc{E}_{t} \mapsto \mathpzc{E}_{t+1}$ and $\mathcal{W}$ is a differentiable warp function based on optical flow and bilinear interpolation implemented using an STN. The self-supervised flavor of this algorithm \cite{evflownet} utilizes corresponding image frames instead of event frames for the loss function but the input is still the stack of event frames. One could utilize the two networks we talked about previously and solve the problem of dodging, however, one would need to run two neural networks for this purpose. Furthermore, this method suffers from a major problem: any unsupervised or self-supervised method can estimate rigid optical flow (optical flow corresponding to the background regions $\mathcal{B}$) accurately but the non-rigid optical flow (optical flow corresponding to the foreground regions $\mathcal{F}$) is not very accurate. This is an artifact because of the number of pixels corresponding to the foreground is often far less than that corresponding to the background, i.e., $\overline{\overline{\mathcal{F}}} \ll \overline{\overline{\mathcal{B}}}$. One would have to train for a lot of iterations to obtain accurate optical flow results on these foreground pixels which runs into the risk of overfitting to the dataset. This defeats the promise of self-supervised or unsupervised formulations.

To solve both the problems of complexity and accuracy, we formulate the problem using a semi-supervised approach to learn segmentation and optical flow at the same time, which we call \textit{EVSegFlowNet}. We call the output of the network \textit{segmentation flow} denoted by $\mathbf{\tilde{\dot{p}}}$ which is defined as follows.
\vspace{-1em}
\begin{equation} \mathbf{\tilde{\dot{p}}_x} =
\mathbf{\dot{x}},\,\, \text{if }\mathds{1}_f\left(\mathbf{x}\right)=1 \text{ and }
\mathbf{\tilde{\dot{p}}_x} = \mathbf{0},\,\, \text{if }\mathds{1}_b\left(\mathbf{x}\right)=1\\
\end{equation}
One could intuit that we can obtain a noisy segmentation for free by simple thresholding on the magnitude of $\mathbf{\tilde{\dot{p}}_x}$. To utilize the network to maximum capacity the input to the network is the ego-motion/odometry based warped event frame such that the background pixels in the two input event frames are almost aligned and the only misalignment comes from the IMOs. This ensures that the network's capacity can be utilized fully for learning sub-pixel accurate optical flow for IMO regions. The input to the EVSegFlowNet is $\mathcal{W}\left(\mathpzc{E}_t, \widetilde{H}_{4Pt}\right)$ and $\mathpzc{E}_{t+1}$. Here, $\widetilde{H}_{4Pt}$ is transformed to $E^F$ before warping.

A complexity analysis of EVSegFlowNet is given in Suppl. Sec. {\color{red}S.VI}. The success of our approach can be seen from the experimental results. The loss function for learning $\mathbf{\tilde{\dot{p}}_x}$ is:
\begin{align}
    \argmin_{\mathbf{\tilde{\dot{p}}_x}}\,\, \mathbb{E}\left( \mathcal{D}\left( \mathcal{W}\left(\mathpzc{E}^{\prime}_t, \mathbf{\tilde{\dot{p}}_x} \right)\circ \mathds{1}_f, \mathpzc{E}_{t+1}\circ \mathds{1}_f\right)\right) +\nonumber\\ \lambda_1 \mathbb{E}\left( \Vert\mathbf{\tilde{\dot{p}}_x}\circ \mathds{1}_b \Vert_1\right) + \lambda_2 \mathbb{E}\left( \Vert\mathbf{\tilde{\dot{p}}_x}\circ \mathds{1}_b \Vert_2^2\right)
    \end{align}

Here, $\lambda_1$ and $\lambda_2$ are regularization parameters. This loss function is essentially the image difference with elastic net like regularization penalty. This penalty makes the network make background flow zero fairly quickly as compared to simple $l_1$ or quadratic penalty whilst being robust to outliers (errors in segmentation mask creation).

Note that all our networks were trained in simulation and directly transfer to the real world without any re-training/fine-tuning. We call our dataset \textit{Moving Object Dataset (MOD)}. Detailed information about the dataset can be found in Suppl. Sec. {\color{red}S.VII}.


\section{Control Policy For Dodging IMOs}
\label{sec:Evasion}
In this section, we present a solution for evading multiple known and/or unknown IMOs. 


\begin{figure}[b!]
    \centering    \includegraphics[width=\columnwidth]{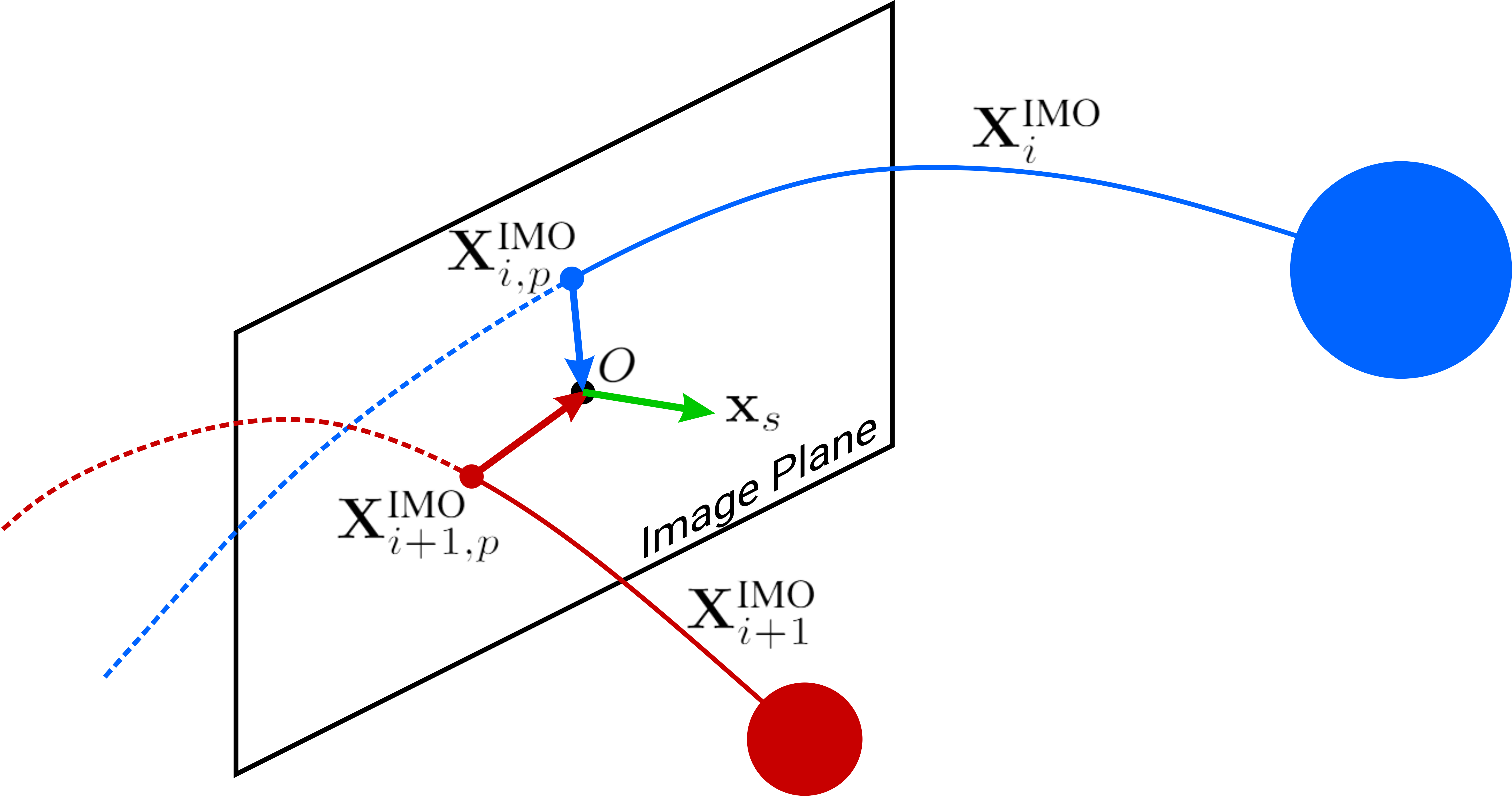}
    \caption{Vectors $\mathbf{X}_{i, p}^{\text{IMO}}$ and $\mathbf{X}_{i+1, p}^{\text{IMO}}$ represent the intersection of the trajectory and the image plane. $\mathbf{x}_{s}$ is the direction of the ``safe'' trajectory. All the vectors are defined with respect to the center of the quadrotor projected on the image plane, $O$. Both of the spheres are of known radii.}
    \label{fig:multipleknown}
\end{figure}

\begin{figure*}[ht!]
    \centering
    \includegraphics[width=0.91\textwidth]{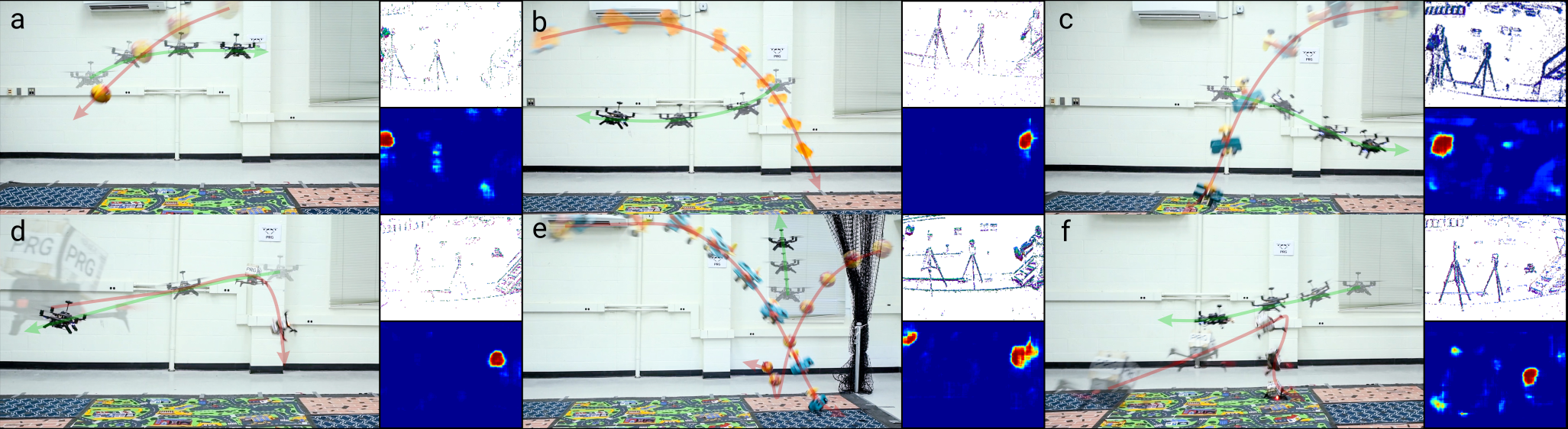}
    \caption{Sequence of images of quadrotor dodging or pursuing of objects. (a)-(d): Dodging a spherical ball, car, airplane and Bebop 2 respectively. (e): Dodging multiple objects simultaneously. (f): Pursuit of Bebop 2 by reversing control policy. Object and quadrotor transparency show progression of time. Red and green arrows indicate object and quadrotor directions respectively. On-set images show front facing event frame (top) and respective segmentation obtained from our network (down).}
    \label{fig:ExperimentImg}
\end{figure*}

Let us consider three different flavors of IMOs: (i) Sphere with known radius $r$, (ii) Unknown shaped objects with known bound on the size and (iii) Unknown objects with no prior knowledge. We tackle each of these cases differently. Knowing the prior information about the geometric nature helps us achieve much more robust results and fine-grain control.
%
We define $\mathcal{F}$ as the projection of all the IMOs on the image plane such that $\mathcal{F} = \bigcup_{\forall i} \mathcal{F}_i$, where $\mathcal{F}_i$ denotes the $i^{\text{th}}$ IMO's image plane projection. Now, let's discuss each flavor of the problem separately in the following subsections.

%

\subsection{Sphere with known radius $r$}
\label{subsec:known-sphere}
\vspace{-0.5em}
Let us first begin with the simplest case, i.e., \textit{a single spherical IMO with known radius $r$}. Evading such an object under no gravitational influence has been tackled and well analyzed by \cite{howfastistoofast}. It is known that the projection of a sphere on the image plane is an ellipse \cite{intro-to-projective-geometry}. For spherical objects under the gravitational influence, we estimate the initial 3D position using the known radius information and then we track the object over a few $\mathpzc{E}$ to obtain the initial 3D velocity. Here, the tracking is done by segmentation on every frame.

Assuming a classical physics model, we predict the future trajectory $\mathbf{X}_i^{\text{IMO}}$ of the sphere when it is only under the influence of gravity. Now, we define the point $\mathbf{X}_{i,p}^{\text{IMO}}$ as the intersection of the trajectory $\mathbf{X}_i^{\text{IMO}}$ and the image plane. For the case of a single spherical IMO, we compute the distance between $\mathbf{X}_{i,p}^{\text{IMO}}$ and the initial position of the quadrotor $O$, denoted by vector $\mathbf{x}_{\text{min}} \in \mathbb{R}^{2\times 1}$. The ``safe'' direction is represented as $\mathbf{x}_s = - \mathbf{x}_\text{min}$. A simple Proportional-Integral-Derivative (PID) controller based on the differential flatness model of the quadrotor is used with high proportional gain for a quick response to move in the direction $\mathbf{x}_s$. The minimum amount of movement is equal to the extended obstacle size (the size of the quadrotor is added to the object size).


Now, let's extend to the evasion of \textit{multiple spherical IMOs}. We assume that while objects are detected, there is no occlusion among different IMOs in the front event camera frame. Then, each object $\mathcal{F}_i$ is clustered using mean shift clustering. For each object $\mathcal{F}_i$, the 3D position and velocity are estimated as before. It is important to note that since all the objects were targeted at the quadrotor, they are bound to intercept the image plane, say at point $\mathbf{X}_{i, p}^\text{IMO}$ (Fig. \ref{fig:multipleknown}). For evasion from multiple objects, we adapt the following approach. First, we find the two objects $m$ and $m+1$ from a consecutive cyclic pair of vectors such that (here $(\hat{\cdot})$ represents a unit vector):
\vspace{-0.1em}
\begin{equation}\argmin_{\mathbf{X}_{i, p}^\text{IMO}, \mathbf{X}_{i+1, p}^\text{IMO}} \left\langle
\hat{\mathbf{X}}_{i, p}^\text{IMO}, \hat{\mathbf{X}}_{i+1, p}^\text{IMO}\right\rangle
\label{eq:argmin-pos}
\end{equation}
In other words, the objects $m$ and $m+1$ forms the largest angle among all the consecutive cyclic pairs. So we deploy a strategy to move the quadrotor in $\mathbf{x}_s$ direction in the image plane such that
\vspace{-0.05em}
\begin{equation} \mathbf{x}_s =
\begin{cases}
- \hat{\mathbf{X}}_{\beta} ,\quad \text{if } \max_{\forall i} \langle\hat{\mathbf{X}}_{\beta}, \mathbf{X}_i\rangle <  \max_{\forall i} \langle-\hat{\mathbf{X}}_{\beta}, \hat{\mathbf{X}}_i\rangle\\
\hat{\mathbf{X}}_{\beta} ,\quad\,\,\,\, \text{otherwise}\\
\end{cases}
\label{eq:strategy}
\end{equation}
\vspace{-0.05em}
where $\mathbf{X}_{\beta} = \hat{\mathbf{X}}_{m, p}^\text{IMO} + \hat{\mathbf{X}}_{m+1, p}^\text{IMO}$


For unknown shaped objects with bound on size, please refer Suppl. Sec. {\color{red}S.VIII}.

\subsection{Unknown objects with no prior knowledge}
Without any prior knowledge about the object, it is geometrically infeasible to obtain the 3D velocity of an IMO using a monocular camera. However, we can predict a possible safe trajectory $\mathbf{x}_s$ depending on the velocity direction of the IMOs on the image plane. We compute the unit vector $ \mathbf{v}_i^\text{IMO}$ in which the IMO is moving by tracking the segmentation mask of the IMO or by computing the mean optical flow direction of the region of interest. For a single unknown IMO, a heuristic is chosen such that the quadrotor moves in the direction perpendicular to the velocity of the IMO on the image plane, i.e., a safe direction for the quadrotor motion which satisfies $\langle \mathbf{x}_{s}, \mathbf{v}_{i}^\text{IMO}\rangle = 0$.


For evasion from multiple objects, we adapt a similar approach as in Sec. \ref{subsec:known-sphere}. First, we find the two objects $m$ and $m+1$ from a consecutive cyclic pair of velocity vectors by replacing $\hat{\mathbf{X}}$ by $\hat{\mathbf{v}}$ in Eq. \ref{eq:argmin-pos}. Now. we deploy a strategy to move the quadrotor in $\mathbf{x}_s$ direction in the image plane by replacing $\hat{\mathbf{X}}$ by $\hat{\mathbf{v}}$ and `$<$' by `$>$' in Eq. \ref{eq:strategy}. Refer to  Suppl. Sec. {\color{red}S.IX} for an extension of our work to pursuit.

\section{Experiments}
\label{sec:Expts}
A detailed description of the hardware and experimental setup is given in Suppl. Sec. {\color{red}S.X}.

\begin{table*}[h!]
\centering
\caption{Quantitative evaluation of different methods for Homography estimation.}
\resizebox{0.95\textwidth}{!}{
\label{tab:HomographyTable}
\begin{tabular}{lllllllllll}
\toprule
\multirow{3}{*}{Method (Loss)} & \multicolumn{4}{c}{\multirow{1}{*}{RMSE$_i$ in px.}} & \multicolumn{4}{c}{\multirow{1}{*}{RMSE$_o$ in px.}} \\
\cline{2-11}\\
 & $\gamma = \pm[0, 5]$ & $\gamma = \pm[6, 10]$ & $\gamma = \pm[11, 15]$ & $\gamma = \pm[16, 20]$ &
 $\gamma =\pm[21, 25]$ & $\gamma = \pm[0, 5]$ & $\gamma = \pm[6, 10]$ & $\gamma = \pm[11, 15]$ & $\gamma = \pm[16, 20]$ & $\gamma =\pm[21, 25]$ \\
\hline\\
Identity & 3.92 $\pm$ 0.83 & 11.40 $\pm$ 0.70 & 18.43 $\pm$ 0.70 & 25.50 $\pm$ 0.70 & 32.55 $\pm$ 0.71 & 3.92 $\pm$ 0.84 & 11.40 $\pm$ 0.70 & 18.44 $\pm$ 0.71 & 25.49 $\pm$ 0.70 & 32.55 $\pm$ 0.71  \\
S & 3.23 $\pm$ 1.13 & 3.90 $\pm$ 1.34 & 5.31 $\pm$ 2.05 & 9.63 $\pm$ 4.57 & 17.65 $\pm$ 7.00 & 4.15 $\pm$ 1.78 & 5.05 $\pm$ 2.19 & 6.99 $\pm$ 3.11 & 11.21 $\pm$ 4.84 & 18.37 $\pm$ 6.61 \\
US$^*$ ($\mathcal{D}_1$) & 2.97 $\pm$ 1.22 & 3.84 $\pm$ 1.61 & 5.99 $\pm$ 2.78 & 11.64 $\pm$ 5.69 & 20.36 $\pm$ 7.68 &  3.92 $\pm$ 1.53 & 5.31 $\pm$ 2.43 & 8.14 $\pm$ 3.86 & 13.63 $\pm$ 5.87 & 21.22 $\pm$ 7.35  \\
US$^*$ ($\mathcal{D}_2$) & 2.48 $\pm$ 0.93 & 3.53 $\pm$ 1.43 & 5.89 $\pm$ 2.70 & 11.74 $\pm$ 5.69 & 20.51 $\pm$ 0.70 & 3.19 $\pm$ 1.26 & 4.86 $\pm$ 2.31  & 7.92 $\pm$ 3.73 & 13.47 $\pm$ 5.71 & 21.22 $\pm$ 7.08  \\
DB + S & 2.73 $\pm$ 1.01 & 3.16 $\pm$ 1.23 & \textbf{4.00 $\pm$ 1.79} & \textbf{6.50 $\pm$ 3.54} & \textbf{12.22 $\pm$ 6.58} & 3.69 $\pm$ 1.51 & \textbf{4.49 $\pm$ 2.10} & \textbf{5.91 $\pm$ 3.16} & \textbf{9.04 $\pm$ 4.90} & \textbf{14.60 $\pm$ 6.95}  \\
DB + US ($\mathcal{D}_1$) & \textbf{2.19 $\pm$ 0.88} & \textbf{3.04 $\pm$ 1.57}  & 4.99 $\pm$ 2.75  & 10.16 $\pm$ 5.54 & 18.62 $\pm$ 7.85 & \textbf{3.08 $\pm$ 1.37} & 4.63 $\pm$ 2.68 & 7.57 $\pm$ 4.30 & 13.16 $\pm$ 6.25 & 21.08 $\pm$ 7.49  \\
DB + US ($\mathcal{D}_2$) & 2.41 $\pm$ 1.06 & 3.30 $\pm$ 1.77 & 5.36 $\pm$ 3.02 & 10.39 $\pm$ 5.78 & 18.77 $\pm$ 8.07 & 3.35 $\pm$ 1.76 & 5.05 $\pm$ 3.03 & 8.11 $\pm$ 4.65 & 13.46 $\pm$ 6.48 & 21.08 $\pm$ 7.81 \\
 \bottomrule
\end{tabular}}\\
\tiny{$^*$ Trained for 100 epochs on supervised and then fine-tuned on unsupervised for 100 more epochs. $\gamma$ denotes the perturbation range in px. for evaluation.}
\end{table*}

\subsection{Experimental Results and Discussion}
\label{sec:ExptResults}
In this paper, we considered the case of navigating through different sets of multiple dynamic obstacles. We dealt with six different evading combinations and one pursuit experiment: (a) Spherical ball with a known radius of 140 mm, (b) car with a bound on the maximum dimension size of 240 mm (with maximum error of $\sim 20\%$ from the original size), (c) airplane with no prior information, (d) Bebop 2 flying at a constant velocity, (e) multiple unknown objects, (f) pursuit of Bebop 2 and (g) low-light dodging experiment. For each evasion case, the objects (Suppl. Fig. {\color{red}S.8}) are directly thrown towards the Aero quadrotor such that a collision would definitely occur if the Aero holds its initial position.
For each case, a total of 30 trials were conducted. Notice that the objects would have hit the quadrotor if it had not moved from its initial position. We achieved a remarkable success rate of $86\%$ in cases (a) and (b), $76\%$ in case (c). Both Parrot Bebop 2 experiments (case (d), (f)) resulted in $83\%$ success rate. Case (e) was carefully performed with synchronized throws between the two objects and resulted about $76\%$ success rate. For the low-light experiment (case (g)), we achieved a success rate of $70\%$. Here success is defined as both a successful detection and evasion for the evade experiments and both a successful detection and collision for the pursuit task. Fig. \ref{fig:ExperimentImg} shows sequences of images for cases (a)-(f) along with sample front facing event frame and segmentation outputs. Vicon plots can be found in Suppl. Fig. {\color{red}S.9}. 

Before the IMO is thrown at the quadrotor, the quadrotor maintains its position (hover) using the differential $X^W$ and $Y^W$ estimates from the EVHomographyNet and $Z^W$ estimates from the sonar.

When the IMO is thrown at the quadrotor, the IMO is detected for five consecutive frames to estimate either the trajectory or image plane velocity and to remove outliers using simple morphological operations. This gives a perception response lag of 60 ms (each consecutive frame pair takes 10 ms for the neural network computation and 2 ms for the post-processing). Finally, the quadrotor moves using the simple PID controller presented before.

Note that, we talked about obtaining both segmentation and optical flow from EVSegFlowNet. This was based on the conceptualization of optical flow being used for other tasks as well. However, if only the dodging task is to be performed, a smaller segmentation network can be used without much loss of accuracy.

\begin{figure}[t!]
    \centering
    \includegraphics[width=\columnwidth]{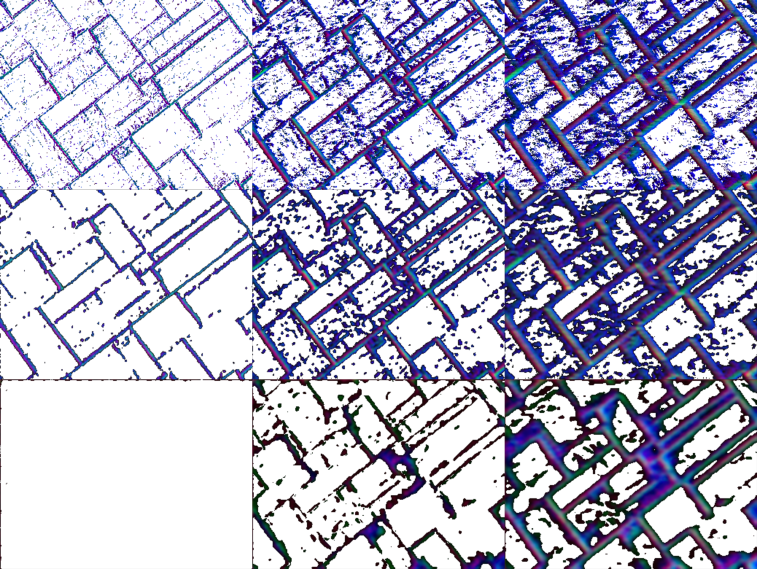}
    \caption{Output of EVDeBlurNet for different integration time and loss functions. Top row: raw event frames, middle row: deblurred event frames with $\mathcal{D}_2$ and bottom row: deblurred event frames with $\mathcal{D}_3$ with $\delta t$. Left to right: $\delta t$ of 1 ms, 5 ms and 10 ms. Notice that only the major contours are preserved and blurred contours are thinned in deblurred outputs.}
    \label{fig:DBOutputs}
\end{figure}

Fig. \ref{fig:DBOutputs} shows the input and output of EVDeBlurNet for losses $\mathcal{D}_2$ and $\mathcal{D}_3$ under $\delta t = \{1, 5, 10\}$ ms. Observe the amount of noise (stray events not associated with strong contours) in the raw images (top row of Fig. \ref{fig:DBOutputs}). The second row shows the output of EVDeBlurNet for $\mathcal{D}_2$ loss. Observe that this works well for smaller integration times but for larger integration times, the amount of denoising and deblurring performance deteriorates. However, $\mathcal{D}_3$ loss which is aimed at outlier rejection is more suppressive of weak contours and hence one can observe that the frame has almost no output for smaller integration times. This has the effect of working well for larger integration times.

\begin{figure}[t!]
    \centering
    \includegraphics[width=\columnwidth]{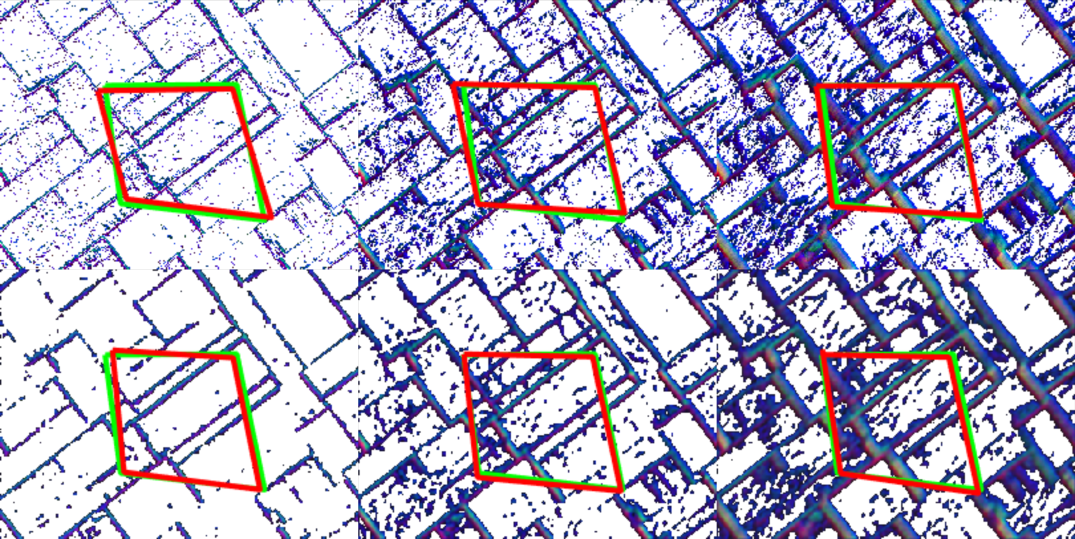}
    \caption{Output of EVHomographyNet for raw and deblurred event frames at different integration times. Green and red color denotes ground truth and predicted $\widetilde{H}_{4Pt}$ respectively.  Top row: raw events frames and bottom row: deblurred event frames. Left to right: $\delta t$ of 1 ms, 5 ms and 10 ms. Notice that the deblurred homography outputs are almost not affected by $\delta t$.}
    \label{fig:DiffHomography}
\end{figure}

Fig. \ref{fig:DiffHomography} shows the output of EVHomographyNet using the supervised loss function on both raw (top row) and deblurred frames (bottom row). Observe that the deblurred homography estimation is more robust to changes in different integration times. The extended version of Table \ref{tab:HomographyTable} is available in Suppl. Sec. {\color{red}S.XI.}) shows the quantitative evaluation of different methods used for training EVHomographyNet. Here, DB represents deblurring using the combination of $\mathcal{D}_2$ and $\mathcal{C}_2$ loss, S and US refer to supervised and unsupervised losses respectively. RMSE$_i$ and RMSE$_o$ denote the average root mean square error  \cite{unsupHomography} in the testing dataset with textures similar to that of the training set, and completely novel textures respectively. RMSE$_o$ quantifies how well the network can generalize to unseen samples. Notice that the supervised flavors of the algorithm work better (lower RMSE$_i$ and RMSE$_o$)  than their respective unsupervised counterparts. We speculate that this is because of the sparsity in data and that the simple image based similarity metrics not being well suited for event frames. We leave crafting a novel loss function for event frames as a potential avenue for future work. Also, notice how deblur variants of the algorithms almost always work better than their respective non-deblurred counterparts highlighting the utility of EVDeblurNet.


\begin{table}[t!]
\centering
\caption{Quantitative evaluation of IMO Segmentation methods.}
\resizebox{\columnwidth}{!}{
\label{tab:SegmentationTable}
\begin{tabular}{llllll}
\toprule
Method & DR$_i$ & DR$_o$ & Run Time & FLOPs & Num. Params\\
(Loss) & in \% & in \% & in ms & in M & in M \\
  \hline\\[-5pt]
  SegNet & 49.0 & 40.4 & 1.5 & 222 & 0.03 \\
  DB + SegNet & 81.5 & 68.7 & 7.5 & 4900 & 2.33\\
  DB + H + SegNet & 83.2 & 69.1 & 10 & 5150 & 3.63\\
  SegFlowNet & 88.3 & 81.9 & 1.5 & 222 & 0.03 \\
  DB + SegFlowNet & 93.3 & 90.1 & 7.5 & 4900 & 2.33 \\
  DB + H + SegFlowNet & 93.4 & 90.7  & 10 & 5150 & 3.63 \\
 \bottomrule
\end{tabular}}
\end{table}

Table \ref{tab:SegmentationTable} shows the quantitative results of different variants of segmentation networks trained using the $\mathcal{D}_2$ loss for SegFlowNet. Also, H denotes the stack of warped $\mathpzc{E}_t$ and $\mathpzc{E}_{t+1}$ using the outputs of the network DB + S in Table \ref{tab:HomographyTable}. Here DR denotes the detection rate and is defined as:
\begin{equation} \text{DR} = \mathbb{E}\left( \sfrac{\overline{\overline{\left(\mathcal{D}\cap \mathcal{G}\right) \circ \mathds{1}_\mathcal{E}}}}{\overline{\overline{\left(\mathcal{G} \circ \mathds{1}_\mathcal{E}\right)}}} \ge \tau \right) \times 100\%\end{equation}

where $\mathcal{G}$ and $\mathcal{D}$ denote the ground truth and predicted masks respectively, and $\mathds{1}_\mathcal{E}$ denotes the presence of an event in either of the input event frames. For our evaluation, we choose $\tau = 0.5$. Notice that using both deblur and homography warping helps improve the results as anticipated. Again, DR$_i$ and DR$_o$ denote testing on trained objects and completely novel objects. As before, deblurring helps generalize much better to novel objects and deblurring with homography warping gives better results showing that the network's learning capacity is utilized better. Also, notice that the improvement in segmentation by warping using homography (last row) is marginal due to the 3D structure of the scene. The network architectures and training details are provided in Suppl. Sec. {\color{red}S.V}.

\section{Conclusions}
\label{sec:Conc}
We presented an AI-based algorithmic design for micro/nano quadrotors, taking into account the knowledge of the system's computation and latency requirements using deep learning. The central conception of our approach is to contrive AI building blocks using shallow neural networks which can be re-purposed. This philosophy was used to develop a method to dodge dynamic obstacles using only a monocular event camera and on-board sensing. To  our  knowledge, this  is  the  first  deep learning  based  solution  to  the  problem  of  dynamic  obstacle avoidance  using  event  cameras  on  a quadrotor. Moreover, our networks are trained in simulation and directly transfer to the real world without fine-tuning or retraining. We also show the generalizability of our navigation stack by extending our work to the pursuit task. As a parting thought, a better similarity scoring metric between event frames or a more robust construction of event frames can improve our results.



\section*{Acknowledgement}
This work was partly funded by the Brin Family Foundation, National Science Foundation under grant BCS 1824198, ONR under grant N00014-17-1-2622, the Northrop Grumman Mission Systems University Research Program. The authors would like to thank NVIDIA for the grant of an Titan-Xp GPU and Intel for the grant of the Aero Platform.

\bibliographystyle{unsrt}
\bibliography{Ref}

\clearpage



\section*{Supplementary material}

# Supplementary Material for
## EVDodgeNet: Deep Dynamic Obstacle Dodging with Event Cameras

Nitin J. Sanket[1], Chethan M. Parameshwara[1], Chahat Deep Singh[1], Ashwin V. Kuruttukulam[1],
Cornelia Fermüller[1], Davide Scaramuzza[2], Yiannis Aloimonos[1]

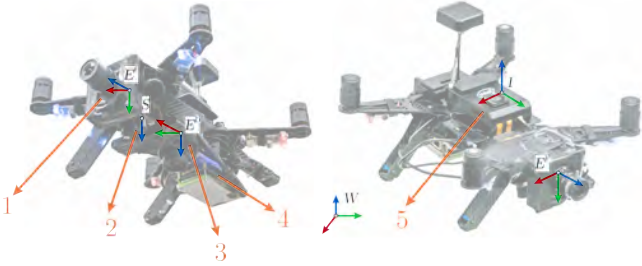

Figure S1. Representation of coordinate frames on the hardware platform used. (1) Front facing DAVIS 240C, (2) down facing sonar on PX4Flow, (3) down facing DAVIS 240B, (4) NVIDIA TX2 CPU+GPU, (5) Intel® Aero Compute board.

## S.I. DEFINITIONS OF COORDINATE FRAMES USED

The letters $I$, $E^F$, $E^D$, $S$ and $W$ denote coordinate frames on the Inertial Measurement Unit (IMU), front facing event camera, down facing event camera, down facing sonar and the world respectively (Fig. S1). All the sensors are assumed to be rigidly attached with the intrinsic and extrinsic calibration between them known. A pinhole camera model is used for the formation of the image. The world point $\mathbf{X}$ gets projected onto the image plane point $\mathbf{x}$. Unless otherwise stated, the points on the image plane are used after undistortion.

## S.II. EVENT FRAME $\mathcal{E}$

A traditional grayscale (global-shutter) camera records frames at a fixed frame rate by integrating the number of photons for the chosen shutter time. This is done for all pixels synchronously. In contrast, an event camera only records the polarity of logarithmic brightness changes *asynchronously at each pixel*. If the brightness at time $t$ of a pixel at location $\mathbf{x}$ is given by $I_{t,\mathbf{x}}$, an event is triggered when:

$$\| \log \left(I_{t+1,\mathbf{x}}\right) - \log \left(I_{t,\mathbf{x}}\right) \|_1 \geq \tau$$

Here $\tau$ is a threshold which will determine if an event is triggered or not. $\tau$ is set at the driver level as a combination of multiple parameters. Each triggered event outputs the following data:

$$\mathbf{e} = \{\mathbf{x}, t, p\}$$

*Nitin J. Sanket and Chethan M. Parameshwara contributed equally to this work. (Corresponding author: Nitin J. Sanket.)*
[1]Perception and Robotics Group, University of Maryland Institute for Advanced Computer Studies, University of Maryland, College Park.
[2]Robotics and Perception Group, Dep. of Informatics, University of Zurich, and Dep. of Neuroinformatics, University of Zurich and ETH Zurich.

where $p = \pm 1$ denotes the sign of the brightness change. The event data unlike an image can vary in data rate and are generally output as a vector of four numbers per triggered event. The data rate is small when the amount of motion and/or scene contrast are small and large when either the motion or the scene contrast is large. This can be beneficial for asynchronous operation, however on a low power digital processor like the one on-board a micro-quadrotor, this can be appalling. To maintain a near constant computational bottle-neck for event processing, we create event frames denoted as $\mathcal{E}$. An event frame is essentially a collection events triggered in a spatio-temporal window starting at $t_0$ and a temporal depth of $\delta t$. The event frame $\mathcal{E}$ is formed as follows.

$$\mathcal{E}\left(\mathbf{x}, \delta t\right)_+ = \sum_{t=t_0}^{t_0+\delta t} \mathbb{1}\left(\mathbf{x}, t, p = +1\right)$$

$$\mathcal{E}\left(\mathbf{x}, \delta t\right)_- = \sum_{t=t_0}^{t_0+\delta t} \mathbb{1}\left(\mathbf{x}, t, p = -1\right)$$

$$\mathcal{E}\left(\mathbf{x}, \delta t\right)_\tau = \left(\sum_{t=t_0}^{t_0+\delta t} \mathbb{1}\left(\mathbf{x}, t, p = \pm 1\right)\right)^{-1} \mathbb{E}\left(t - t_0\right)$$

Here $\mathbb{1}$ is an indicator function which has a value of 1 for an event triggered with polarity of $p$. For $\mathcal{E}_+$ the value of $p$ is +1. Here $\mathbb{E}$ is the expectation/averaging operator. Finally $\mathcal{E}_+$, $\mathcal{E}_-$ and $\mathcal{E}_\tau$ are normalized such that minimum and maximum values are scaled between $[0, 1]$. Essentially $\mathcal{E}_+/\mathcal{E}_-$ captures the per-pixel average number of positive/negative event triggers in the spatio-temporal window spanned between $t_0$ and $t_0 + \delta t$. $\mathcal{E}_\tau$ captures the average trigger time per pixel. This event frame representation is inspired by previous works [1] [2]. The event frame $\mathcal{E}$ is composed by depthwise stacking $\mathcal{E}_+$, $\mathcal{E}_-$ and $\mathcal{E}_\tau$, i.e., $\mathcal{E} = \{\mathcal{E}_+, \mathcal{E}_-, \mathcal{E}_\tau\}$. Using event frames has some pragmatic advantages as compared to processing event by event on the raw event stream.

- The control command can be produced within a constant time bound as event frames are produced at a near constant rate.
- The spatial relationships between event triggers are preserved along with polarity and timing information which is exploited by convolutional neural networks employed in this paper.
- Event frames can be produced in *linear time* in the number of event triggers.

## S.III. EVHomographyNet

A simple and computationally inexpensive way to obtain odometry on a quadrotor is to use a downfacing camera looking at a planar surface. This approximation coupled with data from an IMU and a distance sensor enables high speed "cheap" odometry for navigation. Recently, deep learning approaches have shown more robust homography estimation in traditional images [3], [4]. Inspired by this, we propose the first deep learning based solution to the problem of homography estimation using event cameras which can be run on an embedded computer at reasonably high speeds and good accuracy. Also, the added benefit of using a deep network for homography is that the tradeoff between speed and accuracy could be altered easily (by changing number of parameters). Let us mathematically formulate our problem statement. Let $\mathcal{E}_t$ and $\mathcal{E}_{t+1}$ be the event frames captured at times $t$ and $t+1$, respectively, and $\delta t \ll \Delta t$ where $\Delta t$ is the time difference between the start times of event frame accumulation. In the scenario presented before, the transformation between the two events frames is a homography. This can be written as $\mathbf{x}_{t+1} = \mathbf{H}_t^{t+1}\mathbf{x}_t$, where $\mathbf{x}_{t+1}, \mathbf{x}_t$ represent the homogeneous point correspondences in the two event frames and $\mathbf{H}_t^{t+1}$ is the resulting non-singular $3 \times 3$ homography matrix between the two frames. We adapt the previous works on deep learning based homography estimation [3] [4] for both supervised and unsupervised flavors of deep learning based homography estimation. For the supervised flavor of the algorithm, we generate synthetic homography warped event frames and train them using the following loss function.

$$\operatorname*{argmin}_{\widetilde{H}_{4Pt}} \ \mathbb{E}\left(\|\widetilde{H}_{4Pt} - \hat{H}_{4Pt}\|_2\right) \quad \text{(S1)}$$

Here, $\widetilde{H}_{4Pt}$ and $\hat{H}_{4Pt}$ are the predicted and ground truth 4-point homographies. We refer the readers to [3] for more details.

For the unsupervised version, we adapt the mathematical formulation [4] for TensorDLT and the Spatial Transformer Network (STN) using bilinear interpolation. The final loss function is given as:

$$\operatorname*{argmin}_{\widetilde{H}_{4Pt}} \ \mathbb{E}\left(\mathcal{D}\left(\mathcal{W}\left(\mathcal{E}_t, \widetilde{H}_{4Pt}\right), \mathcal{E}_{t+1}\right)\right) \quad \text{(S2)}$$

where $\mathcal{W}$ is a generic differentiable warp function and can take on different mathematical formulations based on it's second argument (model parameters). In this case, $\mathcal{W}$ contains both the TensorDLT and the STN. As before, $\mathcal{D}$ represents a distance measuring image similarity between two event frames (Refer to the Sec. S.IV for the mathematical formulations of $\mathcal{D}$).

## S.IV. Loss Functions

In this Section, we present the mathematical formulations for variants of the loss functions used in this work.

The two flavors of the heuristic Contrast function $\mathcal{C}$ used in Section II-A of the main paper is inspired by [5] are given below (denoted by $\mathcal{C}_1$ and $\mathcal{C}_2$).

$$\mathcal{C}_1\left(\mathcal{E}\right) = \mathbb{E}\left(\|\operatorname{Var}\left(\nabla \mathcal{E}\right)\|_1\right)$$
$$\operatorname{Var}\left(\mathcal{E}\right) = \mathbb{E}\left(\left(\mathcal{E}\left(\mathbf{x}\right) - \mathbb{E}\left(\mathcal{E}\right)\right)^2\right)$$
$$\mathcal{C}_2\left(\mathcal{E}\right) = \mathbb{E}\left(\|\nabla \mathcal{E}\|_1\right)$$

where $\nabla = \begin{bmatrix} \nabla_x & \nabla_y \end{bmatrix}^T$ is the 2D gradient operator (sobel in our case), Var is the variance operator and $\mathbf{x}$ denotes the pixel location. *The key difference from [5] is that we use the variance operator on the gradients instead on raw values as empirically this gave us better results and was more stable during training.*

Mathematical formulations of the different variants of the distance function $\mathcal{D}$ used in Sections II-A, II-B of the main paper and S.III which measures the similarity between two event frames are given below (denoted as $\mathcal{D}_1$, $\mathcal{D}_2$ and $\mathcal{D}_3$).

$$\mathcal{D}_1\left(\mathcal{E}_1, \mathcal{E}_2\right) = \mathbb{E}\left(\|\mathcal{E}_1 - \mathcal{E}_2\|_1\right)$$
$$\mathcal{D}_2\left(\mathcal{E}_1, \mathcal{E}_2 | \alpha, \epsilon\right) = \mathbb{E}\left(\left(\left(\mathcal{E}_1 - \mathcal{E}_2\right)^2 + \epsilon^2\right)^\alpha\right)$$
$$\mathcal{D}_3\left(\mathcal{E}_1, \mathcal{E}_2 | \alpha, \epsilon, c\right) = \mathbb{E}\left(\frac{b}{d}\left(\left(\frac{\left(\left(\mathcal{E}_1 - \mathcal{E}_2\right)/c\right)^2}{b} + 1\right)^{d/2} - 1\right)\right)$$

$$b = \|2 - \hat{\alpha}\|_1 + \epsilon; \qquad d = \begin{cases} \hat{\alpha} + \epsilon & \text{if } \hat{\alpha} \geq 0 \\ \hat{\alpha} - \epsilon & \text{if } \hat{\alpha} < 0 \end{cases}$$

$$\hat{\alpha}_i = (2 - 2\epsilon_\alpha) \frac{e^{\alpha_i}}{e^{\alpha_i} + 1} \quad \forall i$$

Here, $\mathcal{D}_1$ is the generic $l_1$ photometric loss [6] commonly used for traditional images, $\mathcal{D}_2$ is the Chabonnier loss [7] commonly used for optical flow estimation for traditional images and $\mathcal{D}_3$ is the robust loss function presented in [8]. In $\mathcal{D}_3$, the value of $\alpha$ is output from the network (Refer to S.V for architecture details).

## S.V. Network Details

In this Section, we will present the information on network architecture and training details.

The network architecture is shown in Fig. S2. Notice the simplicity in our network owing our performance to the approach of stacking multiple shallow networks to obtain good performance. *It must be noted that using advanced architectures might lead to better performance.* We leave this as an avenue for future work.

EVDeblurNet was trained for 200 epochs with a learning rate of $10^{-3}$ for 200 epochs with a batch size of 256 for losses using $\mathcal{D}_1$ and $\mathcal{D}_2$ and with a batch size of 32 for losses using $\mathcal{D}_3$. Also, the loss part associated with the contrast is scaled by a factor of 2.0 and the loss part associated with the distance is scaled by a factor of 1.0. This is equivalent to setting $\lambda = 0.33$.

EVSegNet, EVFlowNet, and EVSegFlowNet were trained for 50 epochs with a learning rate of $10^{-4}$ and a batch size of 256. EVHomographyNet was trained for 200 epochs with learning rate $10^{-4}$ and a batch size of 256.

For all the networks, the event frames $\mathcal{E}$ were normalized by dividing each pixel value by 255 and then subtracting by

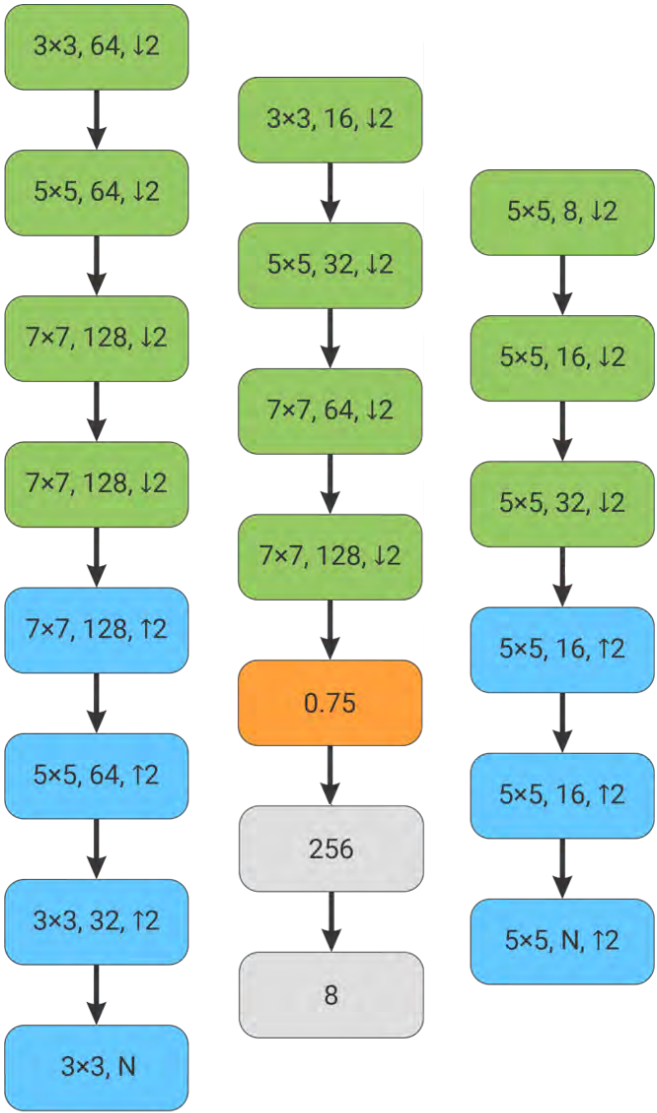

Figure S2. Network Architectures used in the proposed pipeline. Left: EVDeblurNet, Middle: EVHomographyNet and Right: EVSegFlowNet. Green blocks show the convolutional layer with batch normalization and ReLU activation, cyan blocks show deconvolutional layer with batch normalization and ReLU activation and orange blocks show dropout layers. The numbers inside convolutional and deconvolutional layers show kernel size, number of filters and stride factor. The number inside dropout layer shows the dropout fraction. $N$ is 3 and 6 respectively for EVDeblurNet when using losses $\mathcal{D}_1/\mathcal{D}_2$ and $\mathcal{D}_3$. $N$ is 2 and 5 respectively for EVSegFlowNet when using losses $\mathcal{D}_1/\mathcal{D}_2$ and $\mathcal{D}_3$.

0.5 and finally scaling by 2.0 to bound all values between $[-1, 1]$.

For networks using $\mathcal{D}_3$ loss, the values of $\alpha$ are output by the networks last $n$ channels. Here $n$ denotes the number of input channels per image (3 in our case of event frame). $N = 2n$ when using $\mathcal{D}_3$ loss.

### S.VI. Compression Achieved by using EVSegFlowNet

Now, let's analyze the complexity of EVSegFlowNet as compared to a combination of separate segmentation and flow networks we call EVSegNet and EVFlowNet respectively. Let $\mathcal{O}$ denote the complexity measure as the minimum number of neurons to obtain a satisfactory generalization performance on a specific task. We also assume that for smaller and shallow networks complexity scales with number of neurons almost linearly. The complexity for a combination of EVSegNet and EVFlowNet is given by $\mathcal{O}(S)+\mathcal{O}(F)$. Let the complexity of segmentation and flow obtained by EVSegFlowNet be $\mathcal{O}(\widetilde{S})$, $\mathcal{O}(\widetilde{F})$ respectively. A compression/speedup is achieved when $\frac{\mathcal{O}(\widetilde{S}) + \mathcal{O}(\widetilde{F})}{\mathcal{O}(S) + \mathcal{O}(F)} < 1$. Now, because we are only estimating flow for foreground pixels in EVSegFlowNet we have $\frac{\mathcal{O}(\widetilde{F})}{\mathcal{O}(F)} \approx \frac{\overline{\overline{\mathcal{F}}}}{\overline{\overline{\mathcal{B}}}}$. Also, as we mentioned before we get segmentation for free from EVSegFlowNet hence $\mathcal{O}(\widetilde{S}) \approx 0 \ll \mathcal{O}(S)$. This also implies that we achieved good compression/speedup by our formulation as $\frac{\mathcal{O}(\widetilde{S}) + \mathcal{O}(\widetilde{F})}{\mathcal{O}(S) + \mathcal{O}(F)} \ll 1$.

### S.VII. Moving Object Dataset (MOD)

Extensive and growing research on visual (inertial) odometry or SLAM have lead to the development of a large number of datasets. Recent adaption of deep learning to solve these aforementioned problems have fostered the development of large scale datasets (large amount of data). However, most of these datasets are built with the fundamental assumption of static scenes in mind and as a manifestation of which moving or dynamic objects are often not included in these datasets [9]–[11].

To this end, we propose to use synthetic scenes for generating "unlimited" amount of training data with one or more moving objects in the scene. We accomplish this by adapting and proliferating the simulator presented in [10]. To incubate generalization to novel scenes and to utilize the algorithm trained on simulation directly in the real world, we create synthetic moving objects which vary significantly in their texture, shape and trajectory. We also choose random textures for the walls of the 3D room in which objects will move about.

To generate data, we randomize wall textures, objects and object/camera trajectories to obtain seven unique configurations out of which one is exclusively used for test of generalization on more complex structures. Each configuration has a room with three objects moving as shown in Fig. S3. Images are rendered at 1000 frames per second at a resolution of $346 \times 260$ and a field of view of $90°$ for each configuration. Using these images, events are generated following the approach described in [10]. Later event frames $\mathcal{E}$ are generated with three different integration times $\delta t$ of $\{1, 5, 10\}$ ms. Details about the room, lighting and objects are given next.

#### A. 3D room and moving objects

Each room is of size $10 \times 10 \times 5$ m and has random textures on all the walls. These random textures consist of different

3

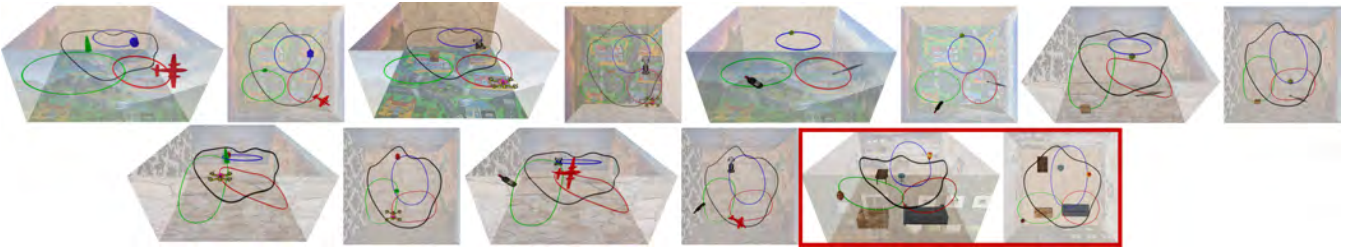

Figure S3. Various Scene setups used for generating data. Red box indicates the scene used for generating out of dataset testing data to evaluate generalization to novel scenes.

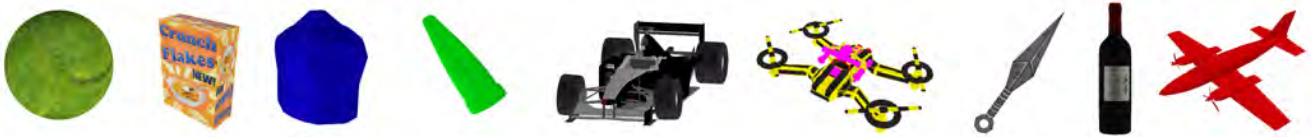

Figure S4. Moving objects used in our simulation environment. Left to right: ball, cereal box, tower, cone, car, drone, kunai, wine bottle and airplane. Notice the variation in texture, color and shape. *Note that the objects are not presented to scale for visual clarity.*

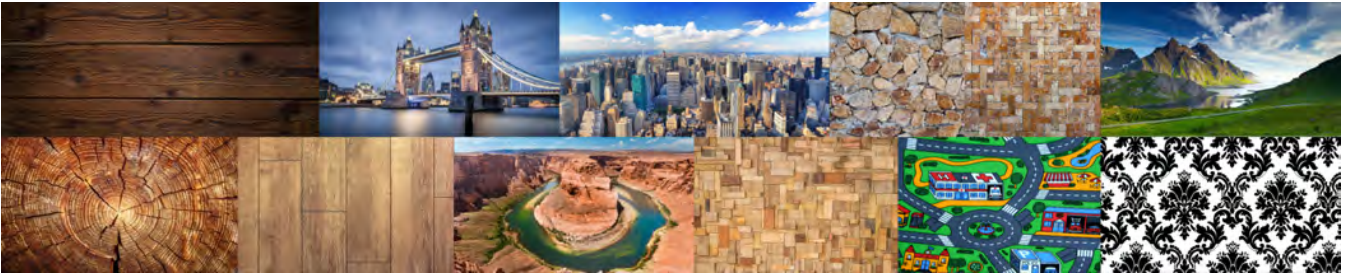

Figure S5. Random textures used in our simulation environment

patterns, colors and shapes. These textures mimic those which occur in real-world indoor and outdoor environments such as skyscrapers, flowers, landscape, bricks, wood, stone and carpet. Each room contains seven light sources inside it for uniform illumination.

The camera is moved inside the 3D room on trajectories such that almost all possible combinations of rotation and translation are obtained. This is aimed at replicating the movement which could be encountered on a real quadrotor.

We have three Independently Moving Objects (IMOs) in each room. Each object is unique in color, shape, texture and size. The objects are chosen to range from simple shapes and textures to complex ones. The objects chosen are ball, cereal box, tower, cone, car, drone, kunai, wine bottle and airplane. The trajectories of the objects are chosen such that many different combinations of relative pose between the camera and the objects are encountered. Also, the objects are moving ten times faster than the camera simulating objects being thrown at a hovering or a slow moving (drifting) quadrotor. The wall textures and moving objects are shown in Figs. S4 and S5 respectively.

### B. Dataset for EVDeblurNet

To learn a simple deblur function, we obtain data from a down facing camera looking at a planar texture and such that no moving objects appear in the frame. The textures for the floor are chosen to replicate the common floor patterns such as wooden flooring, stone, kid's play carpet, and tiles. A total of 15K event frames corresponding to five different textures and integration times $\delta t$ of $\{1, 5, 10\}$ ms are obtained. Random crops of $128 \times 128$ are used to train the network.

### C. Dataset for EVSegNet, EVFlowNet and EVSegFlowNet

In this dataset, the camera follows the same trajectory given in Section S.VII.B to capture the moving objects in the 3D environment. The camera is moving approximately at $0.005$ m per frame and moving objects are ranging from $0.05$ to $0.06$ m per frame. There are some instances where moving objects collide with the camera. We specifically included these scenarios in the dataset so that the learning approaches could learn utilizing both small and large changes in object appearances between consecutive event frames. Average of two objects per frame is captured from the camera (minimum of 0 objects to maximum of 3 objects). Using the six scenarios, 70K event frames are obtained (including data from integration times of $\{1, 5, 10\}$ ms). Event frames from two random integration times per scenario are chosen for training. We obtain 60K images for training and 10K images for testing (we only use 1 ms data not used for training for testing due to mask alignment errors at higher integration times). During training, we use frame skips of

4

Table SI

QUANTITATIVE EVALUATION OF DIFFERENT METHODS FOR HOMOGRAPHY ESTIMATION.

| Method (Loss) | RMSE$_i$ in px. | | | | | RMSE$_o$ in px. | | | | |
|---|---|---|---|---|---|---|---|---|---|---|
| | $\gamma = \pm[0,5]$ | $\gamma = \pm[6,10]$ | $\gamma = \pm[11,15]$ | $\gamma = \pm[16,20]$ | $\gamma = \pm[21,25]$ | $\gamma = \pm[0,5]$ | $\gamma = \pm[6,10]$ | $\gamma = \pm[11,15]$ | $\gamma = \pm[16,20]$ | $\gamma = \pm[21,25]$ |
| Identity | 3.92 ± 0.83 | 11.40 ± 0.70 | 18.43 ± 0.70 | 25.50 ± 0.70 | 32.55 ± 0.71 | 3.92 ± 0.84 | 11.40 ± 0.70 | 18.44 ± 0.71 | 25.49 ± 0.70 | 32.55 ± 0.71 |
| S | 3.23 ± 1.13 | 3.90 ± 1.34 | 5.31 ± 2.05 | 9.63 ± 4.57 | 17.65 ± 7.00 | 4.15 ± 1.78 | 5.05 ± 2.19 | 6.99 ± 3.11 | 11.21 ± 4.84 | 18.37 ± 6.61 |
| US* ($\mathcal{D}_1$) | 2.97 ± 1.22 | 3.84 ± 1.61 | 5.99 ± 2.78 | 11.64 ± 5.69 | 20.36 ± 7.68 | 3.92 ± 1.53 | 5.31 ± 2.43 | 8.14 ± 3.86 | 13.63 ± 5.87 | 21.22 ± 7.35 |
| US* ($\mathcal{D}_2$) | 2.48 ± 0.93 | 3.53 ± 1.43 | 5.89 ± 2.70 | 11.74 ± 5.69 | 20.51 ± 0.70 | 3.19 ± 1.26 | 4.86 ± 2.31 | 7.92 ± 3.73 | 13.47 ± 5.71 | 21.22 ± 7.08 |
| DB + S | 2.73 ± 1.01 | 3.16 ± 1.23 | **4.00 ± 1.79** | 6.50 ± 3.54 | 12.22 ± 6.58 | 3.69 ± 1.51 | 4.49 ± 2.10 | 5.91 ± 3.16 | 9.04 ± 4.90 | 14.60 ± 6.95 |
| DB + US ($\mathcal{D}_1$) | **2.19 ± 0.88** | **3.04 ± 1.57** | 4.99 ± 2.75 | 10.16 ± 5.54 | 18.62 ± 7.85 | **3.08 ± 1.37** | 4.63 ± 2.68 | 7.57 ± 4.30 | 13.16 ± 6.25 | 21.08 ± 7.49 |
| DB + US ($\mathcal{D}_2$) | 2.41 ± 1.06 | 3.30 ± 1.77 | 5.36 ± 3.02 | 10.39 ± 5.78 | 18.77 ± 8.07 | 3.35 ± 1.76 | 5.05 ± 3.03 | 8.11 ± 4.65 | 13.46 ± 6.48 | 21.08 ± 7.81 |
| S$^\mathcal{I}$ | 1.67 ± 0.69 | 2.16 ± 0.92 | **2.92 ± 1.29** | **5.13 ± 2.83** | **11.45 ± 6.08** | 3.02 ± 1.61 | 4.42 ± 2.15 | 6.34 ± 2.84 | 9.38 ± 3.86 | **14.70 ± 5.17** |
| US$^\mathcal{I}$ ($\mathcal{D}_1$) | 1.50 ± 0.59 | 2.16 ± 0.98 | 3.31 ± 1.66 | 6.57 ± 3.85 | 13.45 ± 6.93 | 2.11 ± 0.90 | **3.26 ± 1.46** | 5.34 ± 2.22 | **9.20 ± 3.67** | 15.05 ± 5.27 |
| US$^\mathcal{I}$ ($\mathcal{D}_2$) | **1.49 ± 0.68** | **2.14 ± 1.03** | 3.40 ± 1.69 | 6.91 ± 3.97 | 14.19 ± 6.96 | **2.03 ± 0.92** | 3.31 ± 1.53 | **3.44 ± 2.34** | 9.32 ± 3.60 | 15.53 ± 5.34 |

\* Trained for 100 epochs on supervised and then fine-tuned on unsupervised for 100 more epochs. $\gamma$ denotes the perturbation range in px. for evaluation.

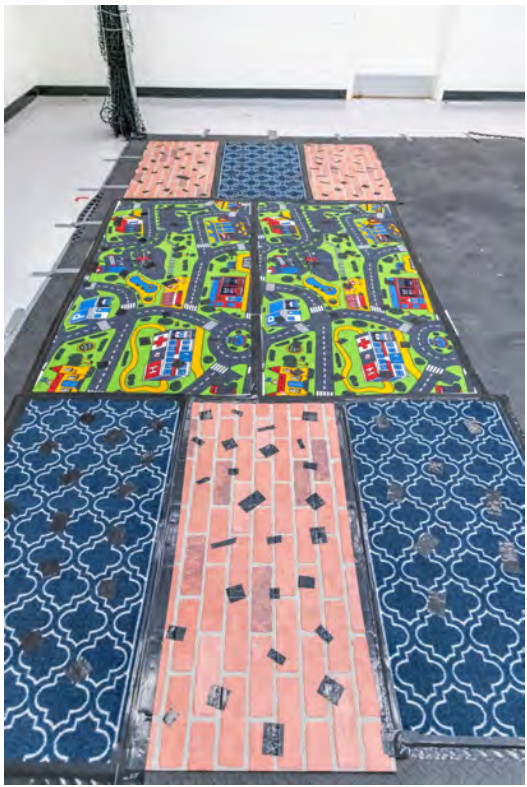

Figure S6. Different textured carpets laid on the ground during real experiments to aid robust homography estimation from EVHomographyNet.

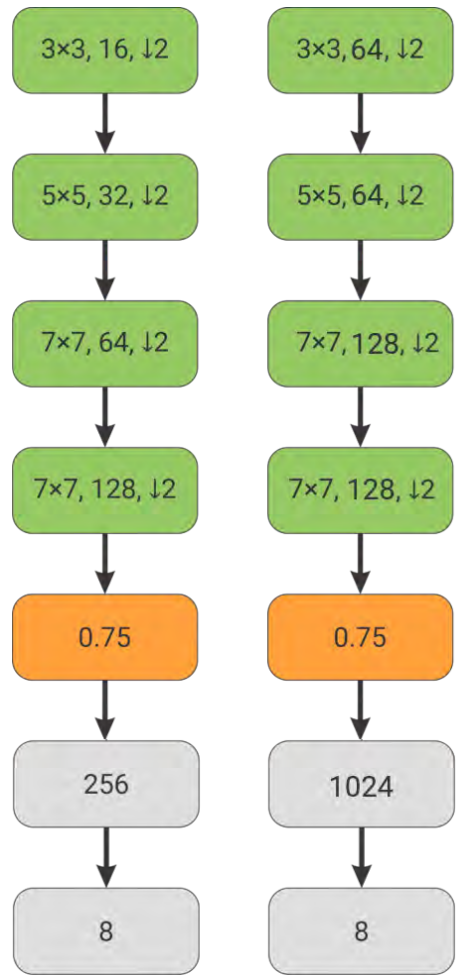

Figure S7. Network Architectures used in Table SII. Left: Homography network used on event frames, Right: Homography network used on image frames. Green blocks show the convolutional layer with batch normalization and ReLU activation, cyan blocks show deconvolutional layer with batch normalization and ReLU activation and orange blocks show dropout layers. The numbers inside convolutional and deconvolutional layers show kernel size, number of filters and stride factor. Notice the similarity in architectures but difference in number of parameters.

Table SII

COMPARSION OF HOMOGRAPHY NETWORK PERFORMANCE FOR EVENT AND RGB FRAMES.

| Input | Run Time in ms | FLOPs in M | Num. Params in M |
|---|---|---|---|
| Event Frame $\mathcal{E}$ | 2.5 | 250 | 1.3 |
| RGB Frame $\mathcal{I}$ | 3.7 | 582 | 9.7 |

one to four to faciliate variable baseline learning of flow and segmentation. We call this test set as "in dataset" testing because the test set though differs significantly in appearance due to integration times still contains the same objects and

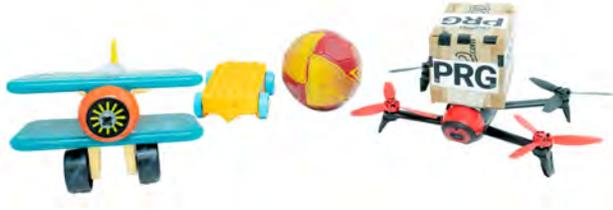

Figure S8. Objects used in experiments. Left to right: Airplane, car, spherical ball and Bebop 2.

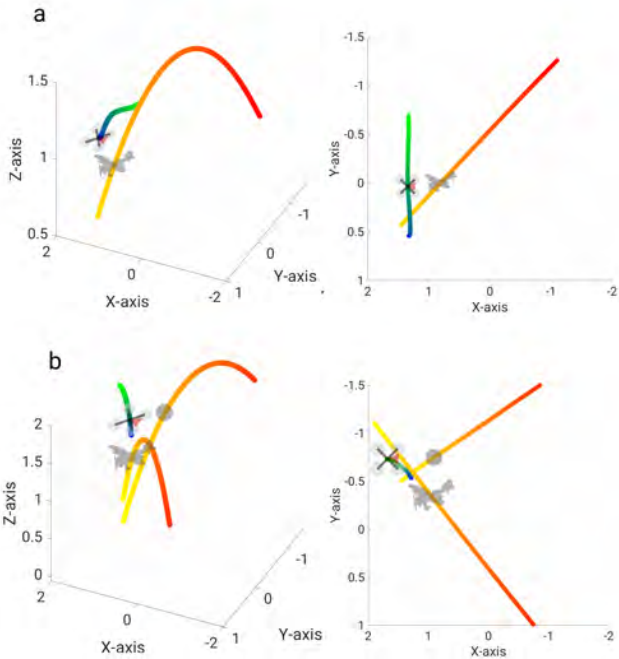

Figure S9. Vicon estimates for the trajectories of the objects and quadrotor. (a) Perspective and top view for single unknown object case, (b) perspective and top view for multiple object case. Object and quadrotor silhouettes are shown to scale. Time progression is shown from red to yellow for objects and blue to green for the quadrotor.

textures as the training set.

For measuring the amount of generalization of our approach, we created a more complex and completely different scenario for testing. This scenario contains immovable 3D objects such as table, chair and a box to 3D room with different textures (different per wall and different from training set textures). The textures on the wall are more realistic depicting a real indoor environment (Refer to the scenario in the red box in Fig. S3). We particularly designed such a scene to highlight that our network is mostly learning from contours and motion information of the objects which is agnostic to scene appearance. Here, we use integration times $\delta t$ of $\{1, 2\}$ ms. We obtain 6K frames for "out of dataset" testing.

### D. Dataset for EVHomographyNet

To train EVHomographyNet, we use the same training set from EVSegFlowNet with the major difference being that here only one frame is used at a time. A random patch of size $128 \times 128$ is obtained from the $346 \times 260$ frame. Then a random perturbation between $\pm\gamma$ is applied to each of the corners. This is used to obtain the homography warped event frame. This approach is exactly the same as given in [3], [4]. For testing "in dataset" the center crops of all the images used for training are chosen and random perturbations of different $\pm\gamma$ are applied. (Refer to Table SI).

For "out of dataset" testing a similar treatment is given to the out of dataset used for evaluating EVSegFlowNet.

### S.VIII. UNKNOWN SHAPED OBJECTS WITH BOUND ON SIZE

Now, consider the case of evading an IMO of an arbitrary shape $\mathcal{S}$. As the projection of $\mathcal{S}$ on the image plane can be either convex or non-convex, we first obtain the convex hull of $\mathcal{S}$ denoted by $\mathcal{H}$. Clearly, an evasive maneuver performed using $\mathcal{H}$ guarantees evasion from the object when the rotation of the IMO with respect to the camera is small.

Next, we find the principal axes of the projection of $\mathcal{H}$ on the image plane. Because we have a bound on size, i.e., we have a bound on the length of the maximum principle axis in 3D, we can evade this object assuming it to be a sphere of this diameter. Note that this method is more conservative than the previous approach constraining the sensing range and latency based on how close the bound is to actual object size.

### S.IX. PURSUIT: A REVERSAL OF EVASION?

The generality of our perception stack for navigation is demonstrated by showing that pursuit can be accomplished using a simple reversal of the control policy for the cases presented in Sec. III.A. and III.B. of the main paper.

Additionally, for an IMO which is self-propelled like a quadrotor, one can perform both pursuit and evade tasks by assuming a linear motion model. Note that here no concept of the agent's intent is used but it can be introduced with an additional neural network for predicting the motion model of the agent (intent) [12]. We leave this for future work.

### S.X. EXPERIMENTAL SETUP

The experiments were conducted in the Autonomy Robotics and Cognition (ARC) lab's indoor flying space at the University of Maryland, College Park. The total flying volume is about $6 \times 5.5 \times 3.5$ m$^3$. A Vicon motion capture system with 8 vantage V8 cameras are used to obtain ground truth at 100 Hz. The objects were either thrown or flown (in-case of the bebop experiment) at the quadrotor during hover or slow flight (simulating slow drift) at speeds ranging from 4.4 ms$^{-1}$ to 6.8 ms$^{-1}$ from a distance ranging from 3.6 m to 5.2 m. To enable robust homography estimation, we laid down carpets of different textures on the ground to obtain strong contours in event frames (Refer to Fig. S6).

We used four different objects in our experiments, (a) a spherical ball of diameter 140 mm, (b) a car of size $185 \times 95 \times 45$ mm (here a bound of 240 mm is used), (c) an airplane of size $270 \times 250 \times 160$ mm (size information not used in experiments), (d) a Bebop 2 of size $330 \times 380 \times 200$ mm.

Also, we used an integration time $\delta t$ of 30 ms for all our experiments.

The proposed framework was tested on a modified Intel® Aero Ready to Fly Drone. The Aero platform was selected for its rugged carbon fiber chassis and integrated flight controller running the PX4 flight stack.

For our experiments, we mounted a front facing DAVIS 240C event camera mated to a 3.3 - 10.5 mm $f$/1.4 lens set at 3.3 mm giving us a diagonal Field Of View (FOV) of 84.5°, a downfacing DAVIS 240B event camera mated to a 4.5 mm $f$/1.4 lens giving us a diagonal FOV of 67.4° and a down facing PX4Flow sensor for altitude measurements. Additionally, we obtain inertial measurments from the IMU on the flight controller. We also mounted an NVIDIA Jetson TX2 GPU to run all the perception and control algorithms on-board (Fig. S1). All the communications happen over serial port or USB. The takeoff weight of the flight setup including the battery is 1400 g with dimensions being $330 \times 290 \times 230$ mm. This gives us a maximum thrust to weight ratio of 1.35.

All the neural networks were prototyped on a PC running Ubuntu 16.04 with and Intel® Core i7 6850K 3.6GHz CPU, an NVIDIA Titan-Xp GPU and 64GB of RAM in Python 2.7 using TensorFlow 1.12. The final code runs on-board the NVIDIA Jetson TX2 running Linux for Tegra® (L4T) 28.1. All the drivers for creating event frames and sensor fusion are written in C++ for efficiency and all the neural network codes run on the TX2's GPU in Python 2.7. We obtain a flight time of about 3 mins.

To enable robust homography estimation, we laid down carpets of different textures on the ground to obtain strong contours in event frames (Refer to Fig. S6).

## S.XI. COMPARISON OF HOMOGRAPHY ESTIMATION USING EVENT AND RGB FRAMES

With 1.3 Million parameters, homography estimation using classical RGB images would not train due to the dearth of number of parameters. The minimum number of parameters required to get reasonable results was 9.7 Million for RGB images and the results are given in Tables SI and SII. These networks were trained on RGB images corresponding to that used for event frame homography networks.

Table SI represents the error comparison of different methods for homography estimation. The last three rows with superscript $\mathcal{I}$ (which were omitted from the original draft due to lack of space) have been included here and they represent the methods trained on RGB images (eighth row denoted by $S^{\mathcal{I}}$) of same resolution as the event frames (fifth row denoted by DB+S). Note that the architecture for the network is the same but the number of parameters is higher (more details about this are given in Table SII and architecture comparisons are shown in Fig. S7). The homography network with the same number of parameters as that used for event frames does not train on RGB frames due to dearth of parameters. Notice that the preformance of best networks for both RGB images and event frames are almost similar despite the dearth in number of parameters in event frame based homography networks, this is because the contour information is more important for homography estimation when the motion is large (large $\gamma$). However, for small perturbations (low $\gamma$), the RGB frame based homography estimation works better due to the dense nature of RGB data which is required for fine alignment. Also, note that the images used for training and testing did not have any motion blur, we expect that in real-world motion blur would degrade the performance of the networks trained on RGB images unless a network to debur RGB frames is used.



\end{document}